\documentclass[times, review, 10pt]{elsarticle}

\usepackage{amssymb}
\usepackage{amsmath}

\usepackage{algorithm}
\usepackage{algorithmicx}
\usepackage{algpseudocode}
\usepackage{bm}
\usepackage[dvipsnames]{xcolor}
\usepackage{colortbl}
\usepackage{tabularx}
\usepackage{adjustbox}
\usepackage{booktabs}
\usepackage{multirow}
\usepackage{caption}
\usepackage{subcaption}
\usepackage{subfig}
\usepackage{pifont}

\journal{Pattern Recognition}

\begin{document}

\begin{frontmatter}



\title{Few-shot Open Relation Extraction with Gaussian Prototype and Adaptive Margin}


\author[inst1]{Tianlin Guo\fnref{fn1}}
\author[inst1]{Lingling Zhang\corref{cor1}\fnref{fn1}}
\author[inst1]{Jiaxin Wang}
\author[inst2]{Yunkuo Lei}
\author[inst2]{Yifei Li}
\author[inst3]{Haofen Wang}
\author[inst2]{Jun Liu}

\affiliation[inst1]{organization={School of Computer Science and Technology and Ministry of Education Key Laboratory of Intelligent Networks and Network Security},
	addressline={Xi’an Jiaotong University}, 
    city={Xi'an},
    postcode={710049}, 
    country={China}}
\affiliation[inst2]{organization={School of Computer Science and Technology and Shaanxi Province Key Laboratory of Big Data Knowledge Engineering},
	addressline={Xi’an Jiaotong University}, 
	city={Xi'an},
	postcode={710049}, 
	country={China}}
\affiliation[inst3]{organization={College of Design and Innovation},
	addressline={Tongji University}, 
	city={Shanghai},
	postcode={200092}, 
	country={China}}

\cortext[cor1]{Corresponding author}
\fntext[fn1]{These authors contributed to the work equllly and should be regarded as co-first authors.}
\fntext[fn2]{\textit{Email addresses}: 1072446164@stu.xjtu.edu.cn (Tianlin Guo),
			 						   zhanglling@xjtu.edu.cn (Lingling Zhang), jiaxinwangg@outlook.com (Jiaxin Wang), 2206113588@stu.xjtu.edu.cn (Yunkuo Lei),
			 						   liyifei619584902@stu.xjtu.edu.cn (Yifei Li), 
			 						   haofen.wang@tongji.edu.cn (Haofen Wang),
			 						   liukeen@xjtu.edu.cn (Jun Liu)}

\begin{abstract}
Few-shot relation extraction with none-of-the-above (FsRE with NOTA) aims at predicting labels in few-shot scenarios with unknown classes. FsRE with NOTA is more challenging than the conventional few-shot relation extraction task, since the boundaries of unknown classes are complex and difficult to learn. Meta-learning based methods, especially prototype-based methods, are the mainstream solutions to this task. They obtain the classification boundary by learning the sample distribution of each class. However, their performance is limited because few-shot overfitting and NOTA boundary confusion lead to misclassification between known and unknown classes. To this end, we propose a novel framework based on Gaussian prototype and adaptive margin named GPAM for FsRE with NOTA, which includes three modules, semi-factual representation, GMM-prototype metric learning and decision boundary learning. The first two modules obtain better representations to solve the few-shot problem through debiased information enhancement and Gaussian space distance measurement. The third module learns more accurate classification boundaries and prototypes through adaptive margin and negative sampling. In the training procedure of GPAM, we use contrastive learning loss to comprehensively consider the effects of range and margin on the classification of known and unknown classes to ensure the model's stability and robustness. Sufficient experiments and ablations on the FewRel dataset show that GPAM surpasses previous prototype methods and achieves state-of-the-art performance.
\end{abstract}

\begin{graphicalabstract}
	\centering
	\includegraphics[width=5cm,height=13cm]{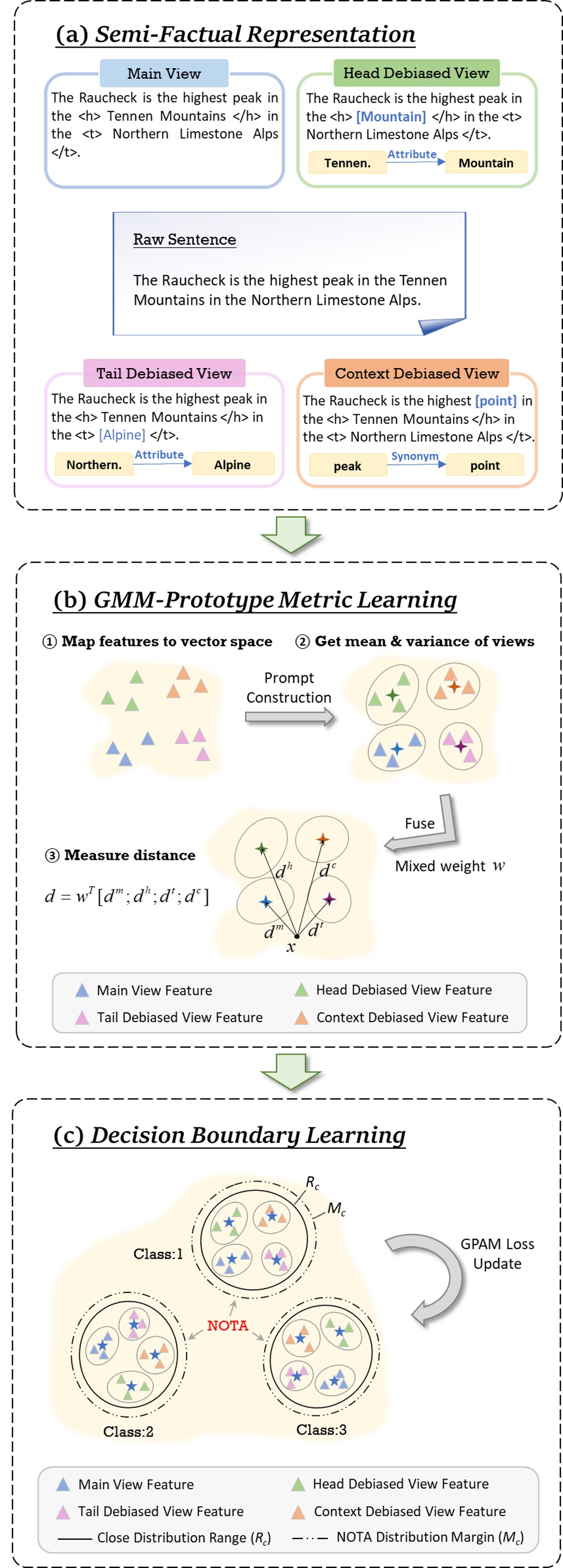}
\end{graphicalabstract}

\begin{highlights}
\item Semi-factual representation helps alleviate the problem of information biases.

\item Gaussian distance metric better captures the distribution of few shots.

\item Adaptive margin gets more accurate boundary decision and prototype.

\item Evaluation on the FewRel dataset achieves state-of-the-art performance.
\end{highlights}


\end{frontmatter}



\section{Introduction}
\label{Introduction}

Relation extraction (RE) is an important task in the field of Natural Language Processing (NLP), which aims to identify and extract semantic relationships between entities from text based on the given list of relations. For the example in Fig.\ref{fig:0}(a), the model should be trained with a large number of examples from the given categories and extract the relation \textit{capital of} when given the entity pair \textit{(Beijing, China)} in the sentence \textit{Beijing is the capital of China}. RE task has been extensively studied in previous work \cite{DBLP:conf/acl/YeL0S22,DBLP:conf/acl/Tian0X22} and existing models can already achieve good classification performance. But for RE task, there is the few-shot issue in the real world, which means that large-scale labeled datasets are difficult to obtain. In addition, sentences that express relations not included in the given set should also be taken into consideration in practical applications, and these unknown classes are called ``none of the above'' (NOTA) \cite{DBLP:conf/emnlp/GaoHZLLSZ19} as the example shown in Fig.\ref{fig:0}(b). One of its characteristics is that the proportion of unknown classes in each task is much larger than that of known classes. For example, for the Few-shot Relation Extraction (FsRE) with NOTA task in Fig.\ref{fig:0}(b), the number of known classes is 2, namely \textit{capital of} and \textit{member of}, and all the remaining classes in the relation set are unknown classes. When given the entity pair \textit{(Maya Airport, Brazzaville)} in the sentence \textit{The airline’s hub is Maya Airport in Brazzaville}, the model should determine that the relation is neither \textit{capital of} nor \textit{member of} in the relation list, and output the result NOTA. To the best of our knowledge, the performance of existing models in dealing with this problem is limited. We summarize that there are two critical challenges in solving this task:

\begin{figure}[t]
	\centering
	\includegraphics[width=1\linewidth]{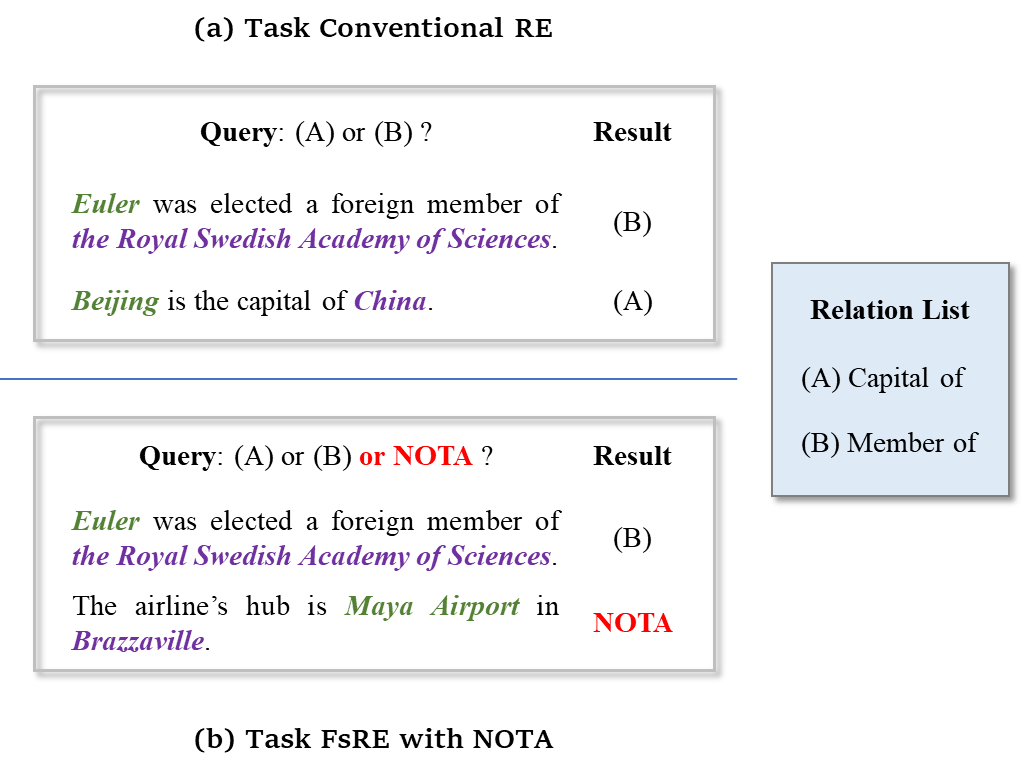}
	\caption{Differences bewteen two tasks RE and FsRE with NOTA. (a) There is a large-scale annotated dataset and all classification results in the query set are within the given set of known classes. (b) There is a few-shot dataset and the query set contains some relations that are not among the known classes and should be classified as NOTA.}
	\label{fig:0}
\end{figure}

\textbf{1) Few-shot Overfitting.} 
Training a classifier for each class can easily lead to overfitting in the case of limited sample supervision. To overcome this issue, many meta-learning approaches, especially prototype-based learning \cite{DBLP:conf/nips/SnellSZ17} ones, have been proposed, which extract the features of each sample and obtain the prototype anchor of each class by averaging or other techniques, and classify the samples based on their distances to these prototypes. However, the traditional prototype-based method has two major flaws. Firstly, directly and simply extracting sample features is incomplete and biased. Relevant studies \cite{DBLP:conf/naacl/WangCZCLLYLH22,DBLP:conf/sigir/HuHZKY23} have demonstrated that head and tail entity information and context information can cause biases in RE model training, which means that over-reliance on entity or context information can cause the model to obtain non-existent relationships when encountering similar entities or contexts. Taking the entity biases as an example, the two sentences \textit{Beijing is the capital of China} and \textit{Beijing is located in China} have the same entity pair (\textit{Beijing, China}) but express different relations, \textit{capital of} and \textit{located in}. Models trained with limited samples are more likely to be confused to obtain wrong relation by these biases. Secondly, the averaging method for prototype computation may fail to accurately represent the distribution of a class. Although Che et al. \cite{DBLP:conf/ijcai/CheA023} introduce the task-specific anchor and combine it with the original class-specific anchor to optimize the generated anchor, this method is more affected by the few-shot scenario and it is difficult to obtain the accurate anchor position using only a few positive samples.

\textbf{2) NOTA Boundary Confusion.} There is a tendency for boundary confusion in the FsRE with NOTA scenario, where instances of the NOTA class may be misclassified as one of the known classes. Only a few conventional models specifically address the NOTA problem, which simply treat NOTA as a single class during training. For example, Liu et al. \cite{DBLP:conf/acl/LiuLHC022} leverage triplet paraphrase to pre-train low-shot relation extraction ability and matches queries and relation labels including NOTA. However, since the semantics of NOTA classes differ in various scenarios, learning their features as a whole fails to capture the correct distribution. This limitation can lead to faults, as the model may incorrectly assign NOTA instances to the closest known class prototypes, rather than recognizing them as belonging to a separate, unknown category. Meanwhile, the few-shot issue will be exacerbated in the presence of the NOTA class, as the model must distinguish between known classes and the unknown class with even fewer data points, increasing the risk of misclassification.

To solve this difficult subject, we propose the framework GPAM, a prototypical learning method using \textbf{G}aussian \textbf{P}rototype and \textbf{A}daptive \textbf{M}argin. As shown in Fig.\ref{fig:1}, our GPAM is mainly composed of three key modules, the semi-facutal representation, the GMM-prototype metric learning and the decision boundary learning module. The first module is to extract better feature representations based on debiased views. The main view and three debiased views are used to deal with the biases caused by the entity and context information shown in Fig.\ref{fig:1}(a). This can reduce the impact of bias information on model training and obtain more accurate prototype representations through augmentation. The second module is to measure the distance between samples and class anchors more accurately. Gaussian distributions with variable mean and variance are used to characterize the distribution of each view, and mixed weights are used to aggregate the four views according to their respective contributions, as is shown in Fig.\ref{fig:1}(b). The first and second modules result in a more accurate distribution and fewer few-shot issues, thereby yielding a precise prototype. The third module is designed to distinguish known and unknown classes and obtain more precise prototype boundaries shown in Fig.\ref{fig:1}(c). To better distinguish known and NOTA classes, the dynamic rather than the fixed margin is introduced and negative sample information is used to optimize the margin range iteratively. For training, contrastive learning strategies are used to make full use of the information of all instances in a query set. This makes the decision boundaries between classes more accurate and reduces NOTA confusion.

Our main contributions can be summarized as follows:

\begin{itemize}
	\setlength{\itemsep}{0pt}
	\item We propose a novel few-shot relation extraction framework GPAM for the challenging NOTA task. This framework alleviates the few-shot issue while ensuring good prototype learning and boundary division effects.
	
	\item We propose a novel prototype distance measurement strategy with Gaussian distribution and introduce semi-factual representation, which alleviates the few-shot overfitting problem.
	
	\item We introduce a new adaptive margin and combine range to optimize decision boundary with contrastive learning loss function. This alleviates the NOTA boundary confusion problem.
	
	\item We conduct experiments on the FewRel dataset to demonstrate the superiority of GPAM. The results show that the model achieves state-of-the-art performance, exhibiting an advantage over previous prototype learning methods.
	
\end{itemize}

\section{Related Work}
\label{Related work}
In this section, we briefly introduce three categories of research relevant to our work, including relation extraction, few-shot relation extraction, and few-shot relation extraction with NOTA.

\subsection{Relation Extraction}
Relation extraction (RE) is a fundamental task in natural language processing (NLP) that aims to identify and classify semantic relationships between entities within a text. Existing researches perform supervised deep learning techniques to gain relations based on large-scale labeled data  \cite{DBLP:conf/acl/ZhouSTQLHX16,DBLP:conf/acl/YeCLL17,DBLP:conf/acl/SoaresFLK19}, which have shown significant improvements in the accuracy. However, data annotation is time-consuming and labor-intensive. Some researchers \cite{DBLP:conf/emnlp/HanYLSL18,DBLP:conf/aaai/ZhouQL023,DBLP:journals/tkde/LiangLLZSG23} use distant supervision methods to align text with external knowledge bases to obtain labels. Liang et al. \cite{DBLP:journals/tkde/LiangLLZSG23} introduce a constraint graph to model the dependencies between labels and shares information between different relation nodes to alleviate long-tail problem. But automatically generated labels may contain noise and depend on the quality of external knowledge bases, affecting model training. With the development of few-shot learning, few-shot relation extraction has gradually become mainstream.

\subsection{Few-shot Relation Extraction}
Providing only a small number of data samples, Han et al. \cite{DBLP:conf/emnlp/HanZYWYLS18} are the first to propose the concept of few-shot relation extraction (FsRE) and introduce the relation extraction dataset FewRel. Recent work on few-shot relation extraction focuses on three main directions, including meta-learning methods, especially prototype-based methods \cite{DBLP:conf/acl/LiuHWC22,DBLP:journals/tnn/XiaoJH23}, parameter optimization learning methods \cite{DBLP:conf/emnlp/GaoHZLLSZ19,DBLP:conf/emnlp/ZhangL22,DBLP:conf/emnlp/0002Q22, DBLP:journals/csl/ZhangHD24} and large language model based methods \cite{DBLP:conf/acl/ZhangG023}. Prototype-based methods aim to obtain the distribution of each category named prototype and assign test samples to the closest one. Xiao et al. \cite{DBLP:journals/tnn/XiaoJH23} propose an adaptive hybrid mechanism based on typical prototype to integrate label information into the features of each category support instance. Parameter optimization learning methods use fine-tuning methods to optimize the parameters of the pre-trained language model to adapt to the RE task. Zhang et al. \cite{DBLP:conf/emnlp/ZhangL22} take relationship descriptions as prompt inputs and randomly discards some to simulate the scenario where support set labels are not visible. Zhang et al. \cite{DBLP:journals/csl/ZhangHD24} utilize prompt-tuning to fine-tune PLM to integrate relationship information and original prototypes. Large language model based methods leverage the vast knowledge and contextual understanding capability to effectively perform tasks in few-shot scenarios. Zhang et al. \cite{DBLP:conf/acl/ZhangG023} utilize an instruction alignment method to fine-tune LLMs and aligns the RE task to the QA task to enhance relation extraction performance. Our work focuses on prototype-based methods and hopes to stimulate the prototype's ability to identify unknown classes, which is called ``none of the above'' (NOTA).

\subsection{Few-shot Relation Extraction with NOTA}
When NOTA is introduced into relation extraction task, an additional challenge arises: accurately distinguishing known classes while precisely identifying and separating the unknown classes. On image classification tasks, there are some studies on FsRE with NOTA. Che et al. \cite{DBLP:conf/mm/SongZL22} treat background class as a pseudo label to train the boundary between known classes and unknown classes. But for text classification tasks, there are much less related studies. BERT-PAIR \cite{DBLP:conf/emnlp/GaoHZLLSZ19} is the first method for this task, which pairs support and query samples for both known and NOTA classes based on a sequence model to calculate the similarity score. Besides, Li et al. \cite{DBLP:conf/emnlp/0002Q22} obtain pseudo labels for pre-training by extracting paraphrase and passes the universal knowledge to the tiny models.

\section{Problem Formalization}
\label{Problem Formalization}
Based on the meta-learning framework, the data is divided into training and validation sets with non-overlapping class labels. Then we construct meta-tasks from the training set, and train the model to achieve optimal performance on the FsRE with NOTA task. The mathematical definition of the task is as follows. For an $N$-way $K$-shot $Q$-query meta task from dataset $D$, we give a formal definition as: $T=\{S,Q^k,Q^u\}$, where $S=\{(x_i, r_i)\}_{i=1}^{NK}$, $Q^k=\{(x_i, r_i)\}_{i=1}^{|Q^k|}$ and $Q^u=\{(x_i, r_i)\}_{i=1}^{|Q^u|}$ denote the support set, the query set from known relations set $C^k$ and the query set from unknown relations set $C^u$, respectively. Models should predict the label $r_i$ corresponding to the query instance $x_i=(h_i,r_i,t_i,c_i)$, and the elements in the quadruple $x_i$ represent the head entity, relationship, tail entity and context respectively. Different from traditional FsRE task, the correct relation label is $r \in \{r_1,r_2,...,r_N,\text{NOTA}\}$ instead of $r \in \{r_1,r_2,...,r_N\}$. 

\section{Methodology}
\label{Methodology}

\begin{figure*}
	\centering
	\includegraphics[width=1\linewidth]{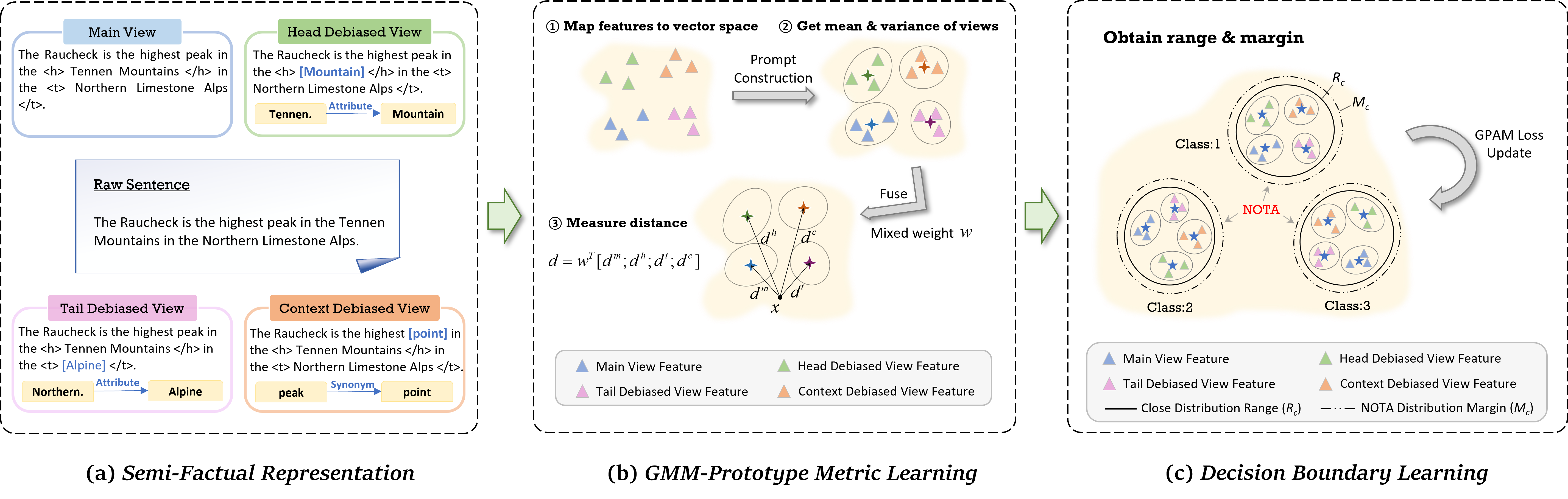}
	\caption{Overview of Our Model GPAM.}
	\label{fig:1}
\end{figure*}

The workflow of our model GPAM is shown in Fig.\ref{fig:1}. Our model consists of three core modules: 1) Semi-Factual Representation, three debiased views are included as semi-factual data derived from the main view to augment the few-shot datasets; 2) GMM-Prototype Metric Learning, the four views' features are fitted to a Gaussian mixture model and obtaining the distance metric; 3) Decision Boundary Learning, prototype range and adaptive margin are used to accurately distinguish various categories and get the decision boundary. We will further introduce these modules as follows.

\subsection{Semi-Factual Representation}
In previous studies \cite{DBLP:conf/nips/SnellSZ17}, prototypes were learned solely from original sentence information, with model performance constrained by biases from entities and contexts. We follow the Semi-Factual Representation (SFR) strategy proposed in our previous work \cite{wang2024learning}, which learns representations through multiple debiased views to mitigate this limitation. As shown in Fig.\ref{fig:1}(a), the details are as follows.

\textbf{Main View.} This view marks the head and tail entity in the raw sentence $\bm{s} = (s_1,s_2,...,s_L)$ with tokens $\langle {h} \rangle$, $\langle {/h} \rangle$ and $\langle {t} \rangle$, $\langle {/t} \rangle$, respectively. We denote the sentence after applying this view as $\bm{s^m} = (s_1^m,s_2^m,...,s_L^m)$.

\textbf{Head and Tail Debiased Views.} These two views replace the head entity $h_i$ and the tail entity $t_i$ with their attribute features. For example, the head entity ``Tennen Mountains'' is replaced by its attribute ``[Mountain]''. The sentences in head debiased view and tail debiased view are denoted by $\bm{s^h} = (s_1^h,s_2^h,...,s_L^h)$ and $\bm{s^t} = (s_1^t,s_2^t,...,s_L^t)$, respectively.

\textbf{Context Debiased View.} This view utilizes WordNet \cite{DBLP:journals/cacm/Miller95} to generate synonyms and randomly replaces 5\% of the words in $\bm{s}$. The resulting sentence is denoted by $\bm{s^c} = (s_1^c,s_2^c,...,s_L^c)$.

We then use BERT \cite{DBLP:conf/naacl/DevlinCLT19} to encode these four views $\bm{s^j} (j = m, h, t, c)$, obtaining feature representations $\bm{z^m}$, $\bm{z^h}$, $\bm{z^t}$, and $\bm{z^c}$, respectively.

\subsection{GMM-Prototype Metric Learning}
The conventional prototype method only calculates the prototype anchor and radius, that is, the prototype is only regarded as a sphere with a certain radius in high-dimensional space \cite{DBLP:conf/ijcai/CheA023}. It only considers the mean of samples, but ignores the variance information that accurately reflects the distribution. This is prone to inaccurate description, especially for samples near the decision boundary. To alleviate this problem, we assume that the relation $r$ follows a mixed Gaussian distribution aggregated from the features of four views and propose a Gaussian Mixture Module (GMM)-based strategy detailed as follows.

First, prompt templates are constructed for four views and inserted into the input embedding sequence as prefix tokens to do prompt-tuning for $K$ shots in a meta-task,
\begin{equation}\label{eq:1}
	input^j = [{prompt}^j] \, [\bm{z}_1^j] [\bm{z}_2^j] \ldots [\bm{z}_K^j]
\end{equation}
where $prompt^j(j=m,h,t,c)$ represents the prompt templates corresponding to various views. Different from traditional high-dimensional space distance estimation strategies, Mahalanobis distance is used instead of Eucilidean distance to better adapt the characteristics of Gaussian distribution, utilizing both the mean and variance information. Next, the mean vector $\bm{\mu}$ and diagonal variance matrix $\text{diag}(\bm{\mathit{v}})$ of the Gaussian space are calculated corresponding to the relation type $r$ and the following formulas are used to get the $\bm{\mu}$, $\bm{v}$ values of the main view and three debiased views in turn:
\begin{equation}\label{eq:2}
	\bm{\mu^j}, \bm{v^j} = \text{Transformer}([{prompt}^j] \, [\bm{z}_1^j] [\bm{z}_2^j] \ldots [\bm{z}_K^j]; \theta)
\end{equation}
where $\theta$ is a learnable parameter which is the same for all views. The four views reflect the features of relation $r$ from different aspects, therefore, the overall feature of $r$ can be obtained by fusing them according to their respective weights. The following function is utilized to calculate the adaptive mixed Gaussian weights $w \in R^4$ of the four views:
\begin{multline}\label{eq:6}
	w = \text{Softmax\_Linear}\left(\text{SelfAttention}\left([\bm{u^m};\bm{v^m}]; [\bm{u^h};\bm{v^h}];\right.\right.\\
	\left.\left.[\bm{u^t};\bm{v^t}]; [\bm{u^c};\bm{v^c}]; \phi_1\right); \phi_2 \right)
\end{multline}
where $w$ is a four-dimensional vector that reflects the relative weights of the four views which differs for different samples, and $\phi_1$ and $\phi_2$ are learnable parameters. Moreover, for any instance $x$ in $Q^k$, $Q^u$, the distance between it and the relation $r$ in the mixed Gaussian prototype space can be measured as:
\begin{equation}\label{eq:7}
	d(x, r) = w^T [d^m(x, r); d^h(x, r); d^t(x, r); d^c(x, r)]
\end{equation}
where for $j=m,h,t,c$:
\begin{equation}\label{eq:8}
	d^j(x, r) = (\bm{z^j} - \bm{u^j})^T \text{diag}(\bm{v^j})(\bm{z^j} - \bm{u^j})
\end{equation}

For the support set $S$, we apply Eq.\eqref{eq:7} and take the average of positive instances to obtain the distance between all samples and candidate relations in the mixed Gaussian space to obtain the distribution feature of each relation category $r_c$. 

\subsection{Decision Boundary Learning}
Since in few-shot scenarios, the distribution of positive samples for a certain category has a key influence on the relation, we introduce the prototype range indicator $R_c$ to show the range of each category with positive samples. However, using only a single indicator $R_c$ to judge the category may cause the problem of sample misjudgment near the decision boundary between known and NOTA classes. In order to alleviate this problem, the adaptive margin of NOTA, $M_c$, is introduced. $M_c$ is another indicator that affects the boundary between positive and negative examples, reflecting the tolerance of the learned prototype to negative examples. We utilize the distances from the positive instances to the relation anchor $r_c$ to obtain $R_c$ and the distances from the negative instances to $r_c$ to obtain $M_c$ as follows:
\begin{equation}\label{eq:9}
	R_c = h_{\tau_1}\left(\left\{d(x^{+}_i, r_c)\right\}^{K}_{i=1}; x^{+}_i \in S\right)
\end{equation}
\begin{equation}\label{eq:10}
	M_c = h_{\tau_2}\left(\left\{d(x^{-}_i, r_c) - R_c\right\}^{(N-1)K}_{i=1} ; x^{-}_i \in S\right)
\end{equation}
where $h(.)$ is the quantile function which follows the principle that most positive instances should be within the prototype range $R_c$ and most negative instances should be outside $R_c + M_c$, $\tau_1$ is a learnable parameter that controls the boundary range, $\tau_2$ is a learnable parameter that controls the tolerance for negative instances, and $x^{+}_i$ and $x^{-}_i$ represent positive and negative instances respectively.

For the query set $Q = Q^k \cup Q^u$, we utilize Eq.\eqref{eq:7} to do the same thing and obtain the distances between instances and each prototype anchor. To determine whether a instance $x$ belongs to relation $r_c$, the classification rules are as follows:
\begin{align*}
	\left\{
	\begin{array}{ll}
		\text{when } d(x, r_c) \leq R_c, & \text{the label of } x \text{ is } r_c, \\
		\text{when } d(x, r_c) > R_c + M_c, & \text{the label of } x \text{ is not } r_c.
	\end{array}
	\right. 
\end{align*}

Besides, if instance $x$ meets the criteria of multiple classes, the instance with the smallest GMM distance $r_c$ will be selected, and if instance $x$ does not belong to any of the relations in the known set, it will belong to the NOTA class. In this way we can accurately predict the labels of the instances in the query set whether they belong to a known class or NOTA class.

In the training procedure, we find that one of the important reasons for the poor performance of prototype learning methods when facing the NOTA problem is that the boundary of NOTA is complex and difficult to accurately describe with a small number of samples. Therefore, we refer to the method of Song \cite{DBLP:conf/mm/SongZL22} to expand some negative instances outside the $R_c + M_c$ region in the background regions by a certain proportion, and these examples are called pseudo negative samples (PNS), which don't belong to any of the classes and serve as negative samples of all classes. According to the research of Wang \cite{DBLP:conf/cvpr/WangHHDS19} et al., since negative instances close to the boundary will have a greater impact on the classification between known and unknown classes, a higher generation probability is assigned to these negative instances. For each meta-task, several points are sampled outside the range of $R_c + M_c$ in the feature vector space as pseudo negative example sets. Then, the probability of selecting a sample in the pseudo-negative example set is calculated by the ratio of its distance from the margin boundary to the closed distribution range according to the following formula:
\begin{equation}\label{eq:11}
	P = \text{Softmax}\left\{\left[\sum_{c=1}^N{\frac{\left\|d(p, r_c) - (R_c + M_c)\right\|}{R_c}}\right]^{-1}\right\}
\end{equation}
where $p$ is a pseudo negative sample generated by this strategy, and is added to the support set $S$ of all classes as a negative instance. After adding pseudo negative samples, we update the value of the range $R_c$ and margin $M_c$ again by Eqs.\eqref{eq:9} and \eqref{eq:10}. 

Then, the contrastive learning strategy is used to optimize our model, and GPAM loss can be formulated as: 
\begin{equation}\label{eq:12}
	\begin{aligned}
		&L_{GPAM} = \frac{1}{N} \sum_{c=1}^N \Biggl\{ \lambda R_c^2\\
		&+ \frac{1}{\alpha} \log \left[ 1 + \sum_{x^{+}_i \in \mathcal{S}_c \cup {Q}^k_c} e^{\alpha (d(x^{+}_i, r_c) - R_c)} \right]\\
		&+ \frac{1}{\beta} \log \left[ 1 + \sum_{x^{-}_i \in \mathcal{S}_c \cup {Q}^k_c \cup {Q}^u_c \cup {P}_c} e^{-\beta (d(x^{-}_i, r_c) - (R_c + M_c))} \right] \Biggr\}
	\end{aligned}
\end{equation}
where $\lambda$ is a hyperparameter that controls the range of the prototype, and $\alpha$ and $\beta$ are adjustable hyperparameters in contrastive learning.

Our loss function can be divided into three components. The initial component focuses on minimizing the range of known classes $R_c$. The middle component ensures that positive examples are positioned as close as possible to the anchors of the known classes. The final component ensures that negative examples are not only distanced from the anchors of the known classes but also placed outside the boundary of $R_c + M_c$. Finally, we summarize our training and optimization process of GPAM in Algorithm 1.

\begin{algorithm}[htbp]
	\caption{The Process of GPAM.}
	\begin{algorithmic}[1]
		\State \textbf{Input:} Support set $ S = \left\{x_i,y_i\right\}^{NK}_{i=1} $; known query set $ Q^k = \left\{x_i,y_i \right\}^{\left|Q^k\right|}_{i=1} $; unknown query set $ Q^u = \left\{x_i,y_i \right\}^{\left|Q^u\right|}_{i=1} $; boundary control parameters $\tau_1$ and $\tau_2$; language model $M$; 
		\State \textbf{Output:} Optimized model parameters of GPAM.
		
		\State \textbf{Procedure} GPAM(Input parameters)
		
		// Semi-Factual Representation Module
		\For{instance i in $S$, $Q^k$ and $Q^u$}
		\State Generate semi-factual views for the input instances.
		\EndFor
		
		\State Initialize variables
		\While{GPAM does not converge}
		\State For instances in $S$, $Q^k$ and $Q^u$, learn the feature of four views by Eq.\eqref{eq:2};
		
		// GMM-Prototype Metric Learning Module
		\State Compute the mixed Gaussian weight of four views by Eq.\eqref{eq:6};
		\State Compute the distance in mixed Gaussian prototype space by Eqs.\eqref{eq:7} and \eqref{eq:8};
		
		// Decision Boundary Learning
		\State Compute the range and margin of prototypes by Eqs.\eqref{eq:9} and \eqref{eq:10};
		\State Expand support set $S$ with pseudo negative instances by Eq.\eqref{eq:11};
		\State Update the range and margin of prototypes by Eqs.\eqref{eq:9} and \eqref{eq:10};
		\State Optimize $M$ by the loss function by Eq.\eqref{eq:12};
		\State Update parameters in the next episode.
		\EndWhile
		
		\Return Model parameters of GPAM
		\State \textbf{End Procedure}
	\end{algorithmic}
\end{algorithm}

\section{Experiment}
\label{Experiment}
In this section, extensive experiments are conducted to compare the proposed GPAM with popular baselines. The detailed framework and learning stage analysis are as follows. 

\subsection{Settings}

\textbf{Datasets.} We perform experiments on the public relation extraction dataset: FewRel \cite{DBLP:conf/emnlp/GaoHZLLSZ19}. FewRel is a benchmark dataset designed for evaluating models in relation classification tasks. It features 100 diverse relation types with annotated examples sourced from Wikipedia, and contains 700 instances for each relation. We use meta-learning methods to randomly extract samples from the FewRel dataset according to different task settings to form meta-task datasets without redundant instances. Different from conventional few-shot learning task, we add NOTA instances to the meta-task datasets according to a certain NOTA rate. 

\textbf{Comparing Methods.} We compare our model GPAM with the following outstanding baselines. These methods can be categorized into three groups. 1) FsRE models with NOTA. Proto-BERT \cite{DBLP:conf/nips/SnellSZ17}: the original prototype network algorithm; BERT-PAIR \cite{DBLP:conf/emnlp/GaoHZLLSZ19}: an approach to measure the	similarity of sentence pairs; MCMN \cite{DBLP:conf/acl/LiuLHC022}: an approach using triplet paraphrase and meta-learning paradigm to do low-shot RE.
2) Those without NOTA. Proto-HATT \cite{DBLP:conf/aaai/GaoH0S19}: a hybrid attention-based prototypical network; MLMAN \cite{DBLP:conf/acl/YeL19}: a multi-level matching and aggregation prototypical network; REGRAB \cite{DBLP:conf/icml/QuGXT20}: a Bayesian meta-learning approach to learn the posterior distribution of the prototype and solve the uncertainty of the prototype vector; CTEG \cite{DBLP:conf/acl-clinicalnlp/WangVB20}: a model to decouple high
co-occurrence relations; HCRP \cite{DBLP:conf/emnlp/Han0L21}: an approach to introduce relation label information and distinguish task difficulty; SimpleFSRE \cite{DBLP:conf/acl/LiuHWC22}: a direct
addition approach that fuses the embedding of relation description to
the prototype representation; SaCon \cite{DBLP:conf/aaai/LuoGHL0CG24}: a framework using diverse viewpoints through instance-label pairs to capture intrinsic textual semantics. 3) Large language models. GPT-4o: an outstanding closed-source large language model developed by OpenAI; GLM-4: an open-source large language model developed by the Tsinghua Zhipu AI team.

\textbf{Training Details.} For GMM-prototype metric learning module, the length of prompt template is set to 100; For decision boundary learning module, the initial values of $\lambda$, $\tau_1$ and $\tau_2$ are set to 0.001, 0.1 and 0.2 respectively, and the ratio of pseudo negative sampling is set to 0.2; In the loss function, the positive impact parameter $\alpha$ is set to 1 and the negative impact parameter $\beta$ is set to 3 to ensure that negative instances have a greater influence in the contrastive learning process. For the training process, we choose the SGD algorithm with learning rate 0.0002, weight decay 0.0001 as the optimizer. 

\subsection{Performance Comparison}
\textbf{Results on FsRE with NOTA task.} Results on FewRel dataset with NOTA are shown in Table \ref{tab:my_label2} and total, known and NOTA are evaluated individually. The observations are as follows:
\begin{itemize}
	
	\item Compared with previous methods, our GPAM clearly achieves state-of-the-art performance on all settings. The total accuracy of our GPAM exceeds the previous best conventional model MCMN, improving by 5.11\%, 4.15\%, 8.19\%, and 7.13\% on four tasks respectively. Benefit from the three designed modules, GPAM achieved good results on the FsRE with NOTA task.
	
	\item GPAM's performance for NOTA class is particularly outstanding under the NOTA rate 0.5 setting. Compared with the previous best performing method MCMN, our GPAM improves the accuracy of NOTA class extraction by 8.85\% $\sim$ 10.30\% at NOTA rate 0.5. The reason is that GMM-based distance metric and adaptive margin are introduced, GPAM has achieved significant advancements in the classification of both known and unknown classes.

	\item As the number of shots $K$ increases, the performance of GPAM becomes higher and more stable. When the number of shots increases from 1 to 5, the accuracy for the NOTA class increases from 84.25 to 93.25 at NOTA rate 0.15, and from 90.75 to 96.10 at NOTA rate 0.5. We can obtain better prototype of categories and more precise decision boundaries with more instances because the performance improvement is reasonable as the shot increases.
	
\end{itemize}

\begin{table}[t]
	\centering
	\caption{Results on FewRel validation dataset with NOTA, known and NOTA are evaluated individually. The optimal values are marked in \textbf{bold}.}
	\label{tab:my_label2}
	\renewcommand{\arraystretch}{1.2}
	\begin{adjustbox}{width=1.0\textwidth}
		\begin{tabular}{ccccccccccccc}
			\toprule
			\multirow{2}{*}{Models}  & \multicolumn{3}{c}{5-way-1-shot 0.15}  & \multicolumn{3}{c}{5-way-5-shot 0.15}  & \multicolumn{3}{c}{5-way-1-shot 0.5}   & \multicolumn{3}{c}{5-way-5-shot 0.5}\\
			\cmidrule(lr){2-4} \cmidrule(lr){5-7} \cmidrule(lr){8-10} \cmidrule(lr){11-13}
			&total  &known  &NOTA  &total  &known  &NOTA   &total  &known  &NOTA   &total  &known  &NOTA\\
			\midrule
			Proto-BERT (NIPS2017) &68.38  &——     &——     &77.25  &——     &——     &40.83  &——     &——     &45.29  &——     &——   \\
			BERT-PAIR (EMNLP2019) &76.42  &77.90  &59.00  &79.62  &83.55  &60.00  &70.63  &72.35  &68.90  &74.42  &76.90  &71.95\\
			MCMN (ACL2022)        &86.47  &87.52  &81.20  &90.72  &91.56  &79.80  &84.29  &85.24  &81.90  &89.70  &91.29  &85.80\\
			GPT-4o                &65.98  &62.06  &85.57  &68.75  &66.25  &81.25  &71.03  &56.29  &85.77  &78.30  &71.06  &85.53\\
			GLM-4                 &90.24  &90.51  &\textbf{88.89}  &91.83  &93.33  &84.31  &87.43  &91.24  &83.62  &84.47  &94.37  &74.56\\
			\textbf{GPAM (Ours)}  &\textbf{91.58}  &\textbf{93.05}  &84.25  &\textbf{94.87}  &\textbf{95.20}  &\textbf{93.25}  &\textbf{92.48}  &\textbf{94.21}  &\textbf{90.75}  &\textbf{95.40}  &\textbf{95.75}  &\textbf{94.10}\\
			\bottomrule
		\end{tabular}%
	\end{adjustbox}
\end{table}

\begin{figure*}
	\centering
	\begin{subfigure}{0.48\linewidth}
		\includegraphics[width=0.98\linewidth]{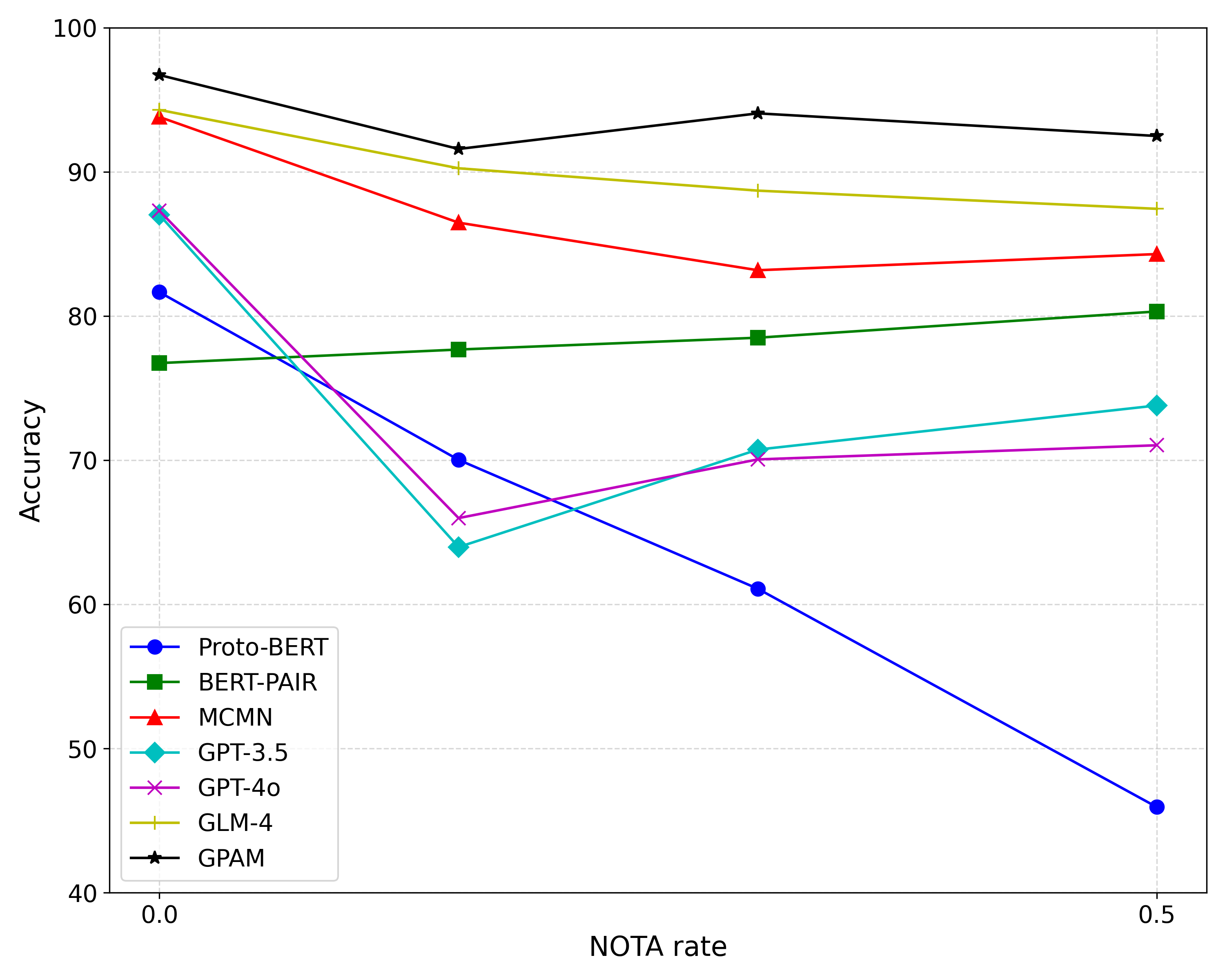}
		\caption{5-way-1-shot}
		\label{fig:subfig1}
	\end{subfigure}
	\hfill
	\begin{subfigure}{0.48\linewidth}
		\includegraphics[width=0.98\linewidth]{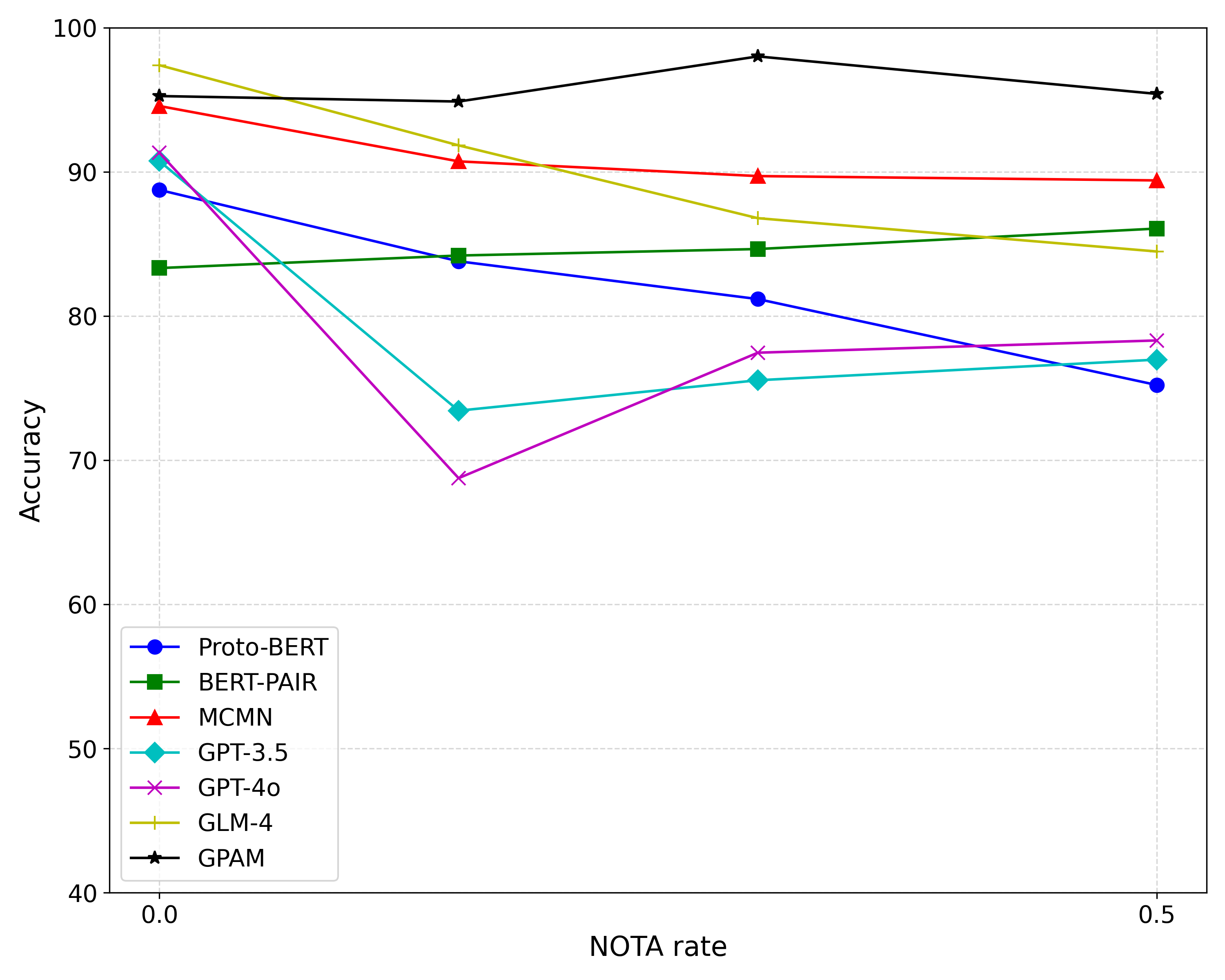}
		\caption{5-way-5-shot}
		\label{fig:subfig2}
	\end{subfigure}
	\caption{Results with increasing NOTA rate from 0 to 50\%.}
	\label{fig:2}
\end{figure*}

Moreover, to compare the performance changes as the NOTA rate improves, we follow the evaluation methods of BERT-PAIR \cite{DBLP:conf/emnlp/GaoHZLLSZ19}, and all models are trained and tested under four different NOTA rates: 0\%, 15\%, 30\%, 50\%. The experimental results are shown in Fig.\ref{fig:2}. GPAM outperforms the compared methods across all NOTA rate settings and maintains better stability. As the NOTA rate increases, the performance of traditional models such as Proto-BERT declines to varying degrees. For the current mainstream large models, GPT-4o's performance drops dramatically after adding NOTA samples compared to  non-NOTA. GLM-4 performs better, but its performance gradually decreases as the NOTA rate increases. It drops by 12.92\% compared to  non-NOTA when NOTA is 0.5. These compared models exhibit significantly weaker discrimination ability for the NOTA class compared to known classes.

\textbf{Results on FsRE task.} We also test the performance on traditional relation extraction task without NOTA shown in Table \ref{tab:my_label1}. It should be noted in advance that since the FewRel dataset test set on Codalab\footnote{https://codalab.org} is no longer open to the public, the validation set is used for all tests. Our GPAM achieves performance slightly inferior to the SOTA model, despite considering the inclusion of unknown class judgments in the zero-NOTA scenario. This shows that our method focuses on tasks including NOTA, but also has acceptable effects on the conventional FsRE task.

\begin{table*}[t]
	\centering
	\caption{Results on FewRel validation / test dataset without NOTA. Note that results of the comparing methods are from papers or Codalab. The optimal and suboptimal values are marked in \textbf{bold} and \underline{underline} respectively.}
	\label{tab:my_label1}
	\renewcommand{\arraystretch}{1.2}
	\begin{adjustbox}{width=1.0\textwidth}
		\begin{tabular}{cllll}
			\toprule
			Models                          &5-way-1-shot  &5-way-5-shot  &10-way-1-shot  &10-way-5-shot\\
			\midrule
			Proto-HATT (AAA2019)            &72.65 / 74.52   &86.15 / 88.40  &60.13 / 62.38  &76.20 / 80.45\\
			MLMAN      (ACL2019)            &75.01 / ——      &87.09 / 90.12  &62.48 / ——     &77.50 / 83.05\\
			Proto-BERT (NIPS2017)           &84.77 / 89.33   &89.54 / 94.13  &76.85 / 83.41  &83.42 / 90.25\\
			BERT-PAIR  (EMNLP2019)          &85.66 / 88.32   &89.48 / 93.22  &76.84 / 80.63  &81.76 / 87.02\\
			REGRAB     (ICML2020)           &87.95 / 90.30   &92.54 / 94.25  &80.26 / 84.09  &86.72 / 89.93\\
			CTEG       (EMNLP2020)          &84.72 / 88.11   &92.52 / 95.25  &76.01 / 81.29  &84.89 / 91.33\\
			HCRP       (EMNLP2021)          &90.90 / 93.76   &93.22 / 95.66  &84.11 / 89.95  &87.79 / 92.10\\
			SimpleFSRE (ACL2022)            &91.29 / 94.42   &94.05 / 96.37  &86.09 / 90.73  &89.68 / 93.47\\
			SimpleFSRE+SaCon (AAAI2024)     &\textbf{98.17} / 97.88   &\textbf{97.98} / 98.12  &\underline{96.21} / 96.65  &\underline{96.46} / 96.50\\
			GPT-4o							&87.29 / ——      &91.32 / ——     &79.35 / ——     &78.05 / ——   \\
			GLM-4							&94.29 / ——      &\underline{97.39} / ——     &\textbf{97.76} / ——     &\textbf{97.02} / ——   \\
			\textbf{GPAM (Ours)}            &\underline{96.71} / ——      &95.25 / ——     &93.85 / ——     &94.75 / ——   \\
			\bottomrule
		\end{tabular}
	\end{adjustbox}
\end{table*}

\subsection{Ablation Study on Semi-Factual Representation}
In order to verify the role of each view and study the impact of each debiased view on the main view, we perform ablation studies on three debiased views. Results are shown in Table \ref{tab:my_label3}. We can make the observations as follows.

\begin{itemize}
	\item The debiased information of all views has a benefit for all task settings. It can be seen that removing any view will cause performance to degrade to varying degrees.
	
	\item The semi-factual representation strategy has a greater improvement in performance in more difficult tasks. Comparing only the main view with the original model performance, we can find that when the nota rate is 0.5, the introduction of debiased views significantly improves the performance by 5.09\% and 6.30\% respectively. 
	
	\item The effect will be better when head and tail views are used together. It can be seen that deleting both the head and tail views at the same time is better than deleting either one. This is because introducing head or tail view alone may lead to incomplete or unbalanced information, and combining the both can get a more reliable representation.
	
\end{itemize}

\subsection{Ablation Study on GMM-Prototype Metric Learning}
To validate the effectiveness of the key strategies in the GMM-prototype metric learning module and to quantify their respective impacts, we construct four variants of GPAM as follows and evaluate their performance on the FewRel dataset shown in Table \ref{tab:my_label3}. The analysis of the results is as follows.

\textbf{Analysis on Guassian Distance:} the variant model that uses Euclidean distance instead of Mahalanobis distance as the metric in Eq.\eqref{eq:8}, that is, treats the prototype as a sphere. Mahalanobis distance shows a more significant improvement in the 5-shot scenario, with increases of 7.70\% and 6.38\%, respectively. This indicates that Mahalanobis distance based on Gaussian distribution is more advantageous than Euclidean distance for prototype construction in scenarios with multiple samples.

\textbf{Analysis on Multi-Prompt:} the variant model that utilizes the same prompt template instead of multi-prompt. We modify Eq.\eqref{eq:2} and set the prompt templates $prompt^j$ of the four views to be the same. Multi-prompt strategy has a more significant effect when the NOTA rate is higher, with improvements of 8.80\% and 7.17\% respectively.

\textbf{Analysis on Mixed Weights:} the variant model that uses the averaging stategy to aggregate four views instead of mixed Gaussian weights. We remove the weight computation formula Eq.\eqref{eq:6}, set the weights of the four views to be the same, and then use Eq.\eqref{eq:7} to obtain the aggregated prototype distance. Mixed weights strategy balances the differences in the importance of different views to certain type of samples, and the effect is more significant in the case of multiple shots. 

\textbf{Analysis on Self-Attention Mechanism:} the variant model that removes the self-attention mechanism used in the prototype. We modify Eq.\eqref{eq:6} by eliminating the self-attention operation, which results of the prototype relying solely on the raw concatenated features without further refinement. Self-Attention has a slight effect compared to other strategies.

Overall, all four strategies have improved the model to a certain extent, and the effects of Mahalanobis distance and multi-prompt are more significant.

\begin{table*}[t]
	\setlength{\aboverulesep}{0pt}
	\setlength{\belowrulesep}{0pt}
	\centering
	\caption{Comparision of ablation results for semi-factual representation, GMM-prototype metric learning, and decision boundary learning. ``$\downarrow$'' represents the change value of accuracy compared with complete GPAM.}
	\label{tab:my_label3}
	\renewcommand{\arraystretch}{1.2}
	\begin{adjustbox}{width=1.0\textwidth}
		\begin{tabular}{lcccc}
			\toprule
			Settings  &5-way-1-shot 0.15  &5-way-5-shot 0.15  &5-way-1-shot 0.5   &5-way-5-shot 0.5\\
			\hline
			\textbf{GPAM}					&\textbf{91.58}  &\textbf{94.87}  &\textbf{92.48}  &\textbf{95.40}\\
			\midrule
			\multicolumn{2}{l}{\textit{\textbf{Semi-Factual Representation}}}   &&&\\
			- Main Only                 &89.96 ($\downarrow$-1.60)  &91.17 ($\downarrow$-3.70)  &87.39 ($\downarrow$-5.09)  &89.10 ($\downarrow$-6.30)\\
			- w/o Head Debiased         &90.75 ($\downarrow$-0.83)  &91.04 ($\downarrow$-3.83)  &89.40 ($\downarrow$-3.08)  &91.45 ($\downarrow$-3.95)\\
			- w/o Tail Debiased         &90.21 ($\downarrow$-1.37)  &90.25 ($\downarrow$-4.62)  &88.83 ($\downarrow$-3.65)  &91.15 ($\downarrow$-4.25)\\
			- w/o Context Debiased      &89.99 ($\downarrow$-1.59)  &92.67 ($\downarrow$-2.20)  &91.00 ($\downarrow$-1.48)  &91.08 ($\downarrow$-4.32)\\
			- w/o Head\&Tail Debiased   &90.29 ($\downarrow$-1.29)  &91.87 ($\downarrow$-3.00)  &90.50 ($\downarrow$-1.98)  &92.20 ($\downarrow$-3.20)\\
			\midrule
			\midrule
			\multicolumn{2}{l}{\textit{\textbf{GMM-Prototype Metric Learning}}}   &&&\\
			- w/o Guassian Distance      	&87.04 ($\downarrow$-4.54)  &87.17 ($\downarrow$-7.70)  &88.20 ($\downarrow$-4.28)  &89.02 ($\downarrow$-6.38)\\
			- w/o Multi-Prompt              &85.87 ($\downarrow$-5.71)  &90.54 ($\downarrow$-4.33)  &83.68 ($\downarrow$-8.80)  &88.23 ($\downarrow$-7.17)\\
			- w/o Mixed Weights     		&90.16 ($\downarrow$-1.42)  &89.42 ($\downarrow$-5.45)  &86.70 ($\downarrow$-5.78)  &90.83 ($\downarrow$-4.57)\\
			- w/o Self-Attention    		&91.25 ($\downarrow$-0.33)  &93.83 ($\downarrow$-1.04)  &90.73 ($\downarrow$-1.75)  &92.65 ($\downarrow$-2.75)\\
			\midrule
			\midrule
			\multicolumn{2}{l}{\textit{\textbf{Decision Boundary Learning}}}   	&&&\\
			- w/o Margin   	     		    &89.96 ($\downarrow$-1.62)  &86.43 ($\downarrow$-8.44)  &86.23 ($\downarrow$-6.25)  &87.42 ($\downarrow$-7.98)\\
			- w/o Adaptive Margin  			&85.71 ($\downarrow$-5.87)  &91.50 ($\downarrow$-3.37)  &89.77 ($\downarrow$-2.71)  &89.90 ($\downarrow$-5.50)\\
			- w/o PNS             			&90.46 ($\downarrow$-1.12)  &91.58 ($\downarrow$-3.29)  &88.94 ($\downarrow$-3.54)  &92.00 ($\downarrow$-3.40)\\
			\bottomrule
		\end{tabular}%
	\end{adjustbox}
\end{table*}

\subsection{Ablation Study on Decision Boundary Learning}
As shown in Table \ref{tab:my_label3}, we also conduct an ablation study on decision boundary learning module, analyzing the impact of the presence or absence of key strategies as follows.

\textbf{Analysis on Class Margin $M_c$.} In order to study the impact of margin presence and dynamics on performance, we construct two variants of GPAM: 1) The variant model that utilizes only range indicator $R_c$ without margin $M_c$ to classify unknown classes. We achieve this by setting $M_c$ to zero and remove this item in the loss function Eq.\eqref{eq:12}. It can be seen that in the case of a relatively simple task 5-way-1-shot 0.15, the increase of the margin strategy is small, but in the other three complex tasks, the presence or absence of margin has a great impact, with an increase of more than 6\%. This indicates that measuring the boundary solely with the prototype range is inaccurate, and the margin is crucial. 2) The variant model that utilizes fixed margin instead of adaptive margin which changes with negative instances of the prototype. We achieve this by changing $M_c$ in the margin computation formula Eq.\eqref{eq:10} and the loss function Eq.\eqref{eq:12} to a fixed value. Taking 5-way-1-shot 0.15 task as an example, using a fixed margin reduces the performance by 5.87\%, which is even more significant than simply removing the margin. One possible reason is that when there are fewer shots, using the same $M_c$ for different categories will lead to inaccurate boundary ranges.

\begin{figure}[t]
	\centering
	\includegraphics[width=1\linewidth]{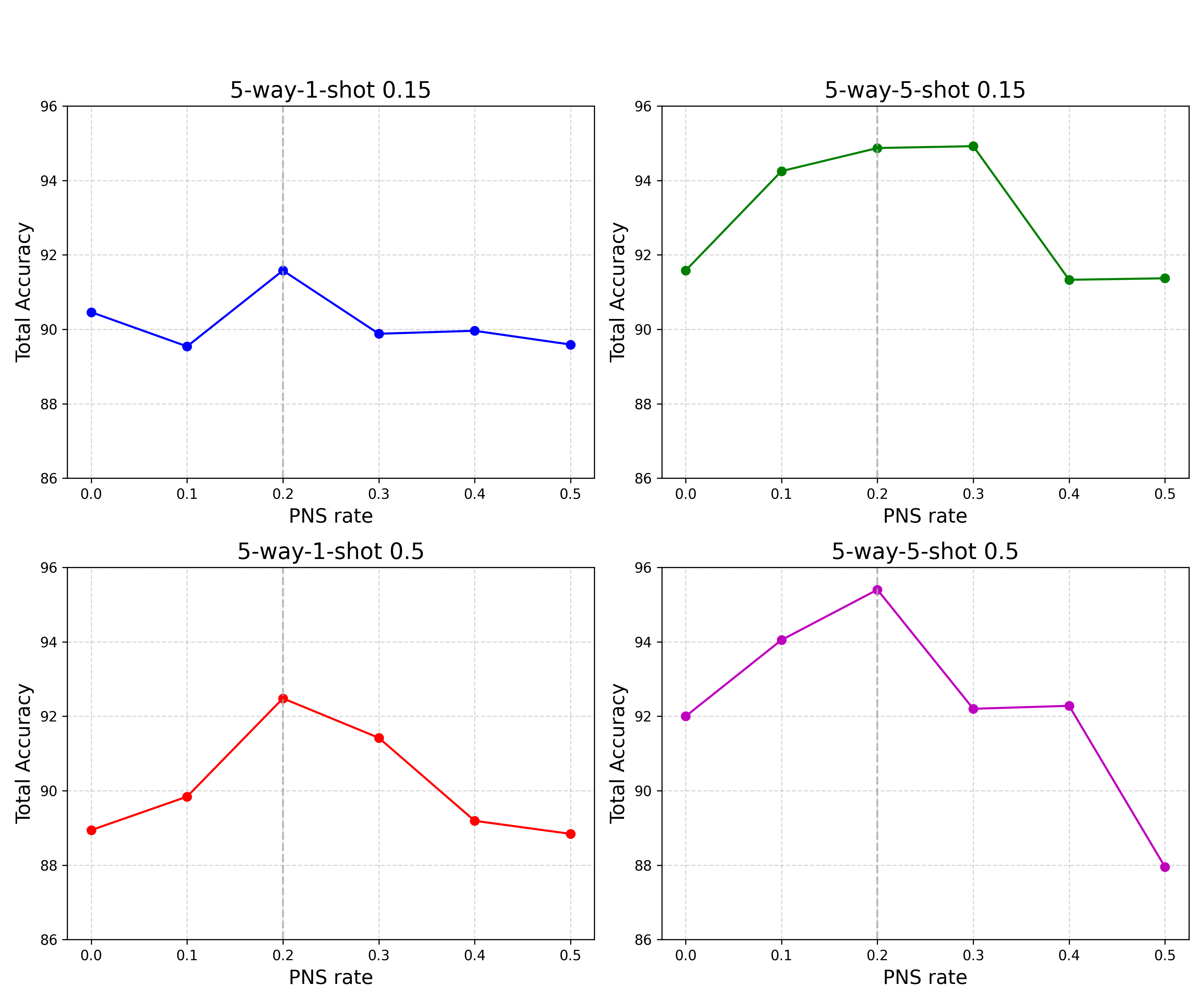}
	\caption{The performance variation curves for different negative sampling rates under four task settings.}
	\label{fig:3}
\end{figure}

\textbf{Analysis on Pseudo Negative Sampling (PNS).} We construct a variant model with no pseudo negative samples in the train dataset. It can be seen that the PNS strategy shows a modest improvement of only 1.12\% for the 5-way-1-shot task with the NOTA rate of 0.15, while it achieved over 3\% improvement for the other three tasks. We analyze the reason for the results and infer that for the 5-way-1-shot 0.15 task, adding pseudo negative examples expands the boundary range of prototype. As a result, more NOTA classes are misclassified as known classes, even though the overall performance is improved. In order to obtain the optimal negative sampling rate, we conduct experiments and plot the performance of different negative sampling rates as shown in Figure \ref{fig:3}. The performance is optimal in most cases when the negative sampling rate is 0.2, and the effect drops significantly when the value is 0.4 or higher.

\definecolor{customgreen}{rgb}{0.0, 0.5, 0.0}
\definecolor{custompurple}{rgb}{0.5, 0.0, 0.5}
\definecolor{darkgreen}{rgb}{0.0, 0.0, 0.7}
\definecolor{darkred}{rgb}{0.7, 0.0, 0.0}

\begin{table*}[t]
	\centering
	\caption{Case study on 5-way-1-shot 0.5 meta tasks. The support set is omitted, and only the query set instances and output results are shown. Each row in the table represents the query, ground-truth, and results of different models in a meta-task.}
	\label{tab:my_label7}
	\begin{adjustbox}{width=1.0\textwidth}
		\begin{tabular}{cp{7cm}cccccc}
			\toprule
			\multicolumn{2}{c}{Query} & Ground-Truth & BERT-PAIR & GPAM & $\text{GPAM}^{\#1}$ & $\text{GPAM}^{\#2}$ & $\text{GPAM}^{\#3}$\\
			\midrule
			\multirow{12}{*}{Known}
			&(1) As surety for the accord, Lambert pledged to marry \textcolor{custompurple}{\textbf{Gisela}}, \textcolor{customgreen}{\textbf{Berengar}}'s daughter.
			&\multirow{2}{*}{P25 mother}  &\multirow{2}{*}{P25 \textcolor{darkgreen}{\ding{51}}}  &\multirow{2}{*}{P25 \textcolor{darkgreen}{\ding{51}}}  &\multirow{2}{*}{P25 \textcolor{darkgreen}{\ding{51}}}  &\multirow{2}{*}{P25 \textcolor{darkgreen}{\ding{51}}}  &\multirow{2}{*}{P25 \textcolor{darkgreen}{\ding{51}}}\\
			\cline{2-8}
			&(2) Led Zeppelin, used the mobile studio to record material for the albums \textcolor{customgreen}{\textbf{`Physical Graffiti''}} and \textcolor{custompurple}{\textbf{Houses of the Holy''}}. 
			&\multirow{3}{*}{P155 follows}  &\multirow{3}{*}{P155 \textcolor{darkgreen}{\ding{51}}}  &\multirow{3}{*}{P155 \textcolor{darkgreen}{\ding{51}}}  &\multirow{3}{*}{P155 \textcolor{darkgreen}{\ding{51}}}  &\multirow{3}{*}{P155 \textcolor{darkgreen}{\ding{51}}}  &\multirow{3}{*}{P155 \textcolor{darkgreen}{\ding{51}}}\\
			\cline{2-8}
			&(3) Minsk Zoo is located in a southeast part of \textcolor{customgreen}{\textbf{Minsk}} near \textcolor{custompurple}{\textbf{Svislach River}}. 
			&\multirow{2}{*}{P206 located in body of water}  &\multirow{2}{*}{P206 \textcolor{darkgreen}{\ding{51}}}  &\multirow{2}{*}{P206 \textcolor{darkgreen}{\ding{51}}} &\multirow{2}{*}{P206 \textcolor{darkgreen}{\ding{51}}}  &\multirow{2}{*}{P206 \textcolor{darkgreen}{\ding{51}}}  &\multirow{2}{*}{P206 \textcolor{darkgreen}{\ding{51}}}\\
			\cline{2-8}
			&(4) Julius Peppers held out of team drills, and \textcolor{customgreen}{\textbf{Chauncey Davis}} was called to take first team reps at \textcolor{custompurple}{\textbf{defensive end}}. 
			&\multirow{3}{*}{P413 position played on team}  &\multirow{3}{*}{P206 \textcolor{darkred}{\ding{55}}}  &\multirow{3}{*}{P413 \textcolor{darkgreen}{\ding{51}}}  &\multirow{3}{*}{P413 \textcolor{darkgreen}{\ding{51}}}  &\multirow{3}{*}{P206 \textcolor{darkred}{\ding{55}}}  &\multirow{3}{*}{P413 \textcolor{darkgreen}{\ding{51}}}\\
			\cline{2-8}
			&(5) Magas was unofficially proclaimed as the \textcolor{customgreen}{\textbf{godfather}} of Serbian \textcolor{custompurple}{\textbf{organized crime}} at the time. 
			&\multirow{2}{*}{P921 main subject}  &\multirow{2}{*}{NOTA \textcolor{darkred}{\ding{55}}}  &\multirow{2}{*}{P921 \textcolor{darkgreen}{\ding{51}}}  &\multirow{2}{*}{NOTA \textcolor{darkred}{\ding{55}}}  &\multirow{2}{*}{P921 \textcolor{darkgreen}{\ding{51}}}  &\multirow{2}{*}{NOTA \textcolor{darkred}{\ding{55}}}\\
			\midrule
			\midrule
			\multirow{19}{*}{Unknown}
			&(6) The Mary Hill Bypass, officially known as Highway 7B, runs adjacent to the \textcolor{custompurple}{\textbf{Fraser River}} from the Pitt River Bridge on the east to the \textcolor{customgreen}{\textbf{Port Mann Bridge}} on the west. 
			&\multirow{4}{*}{\textcolor{red}{NOTA} (P177)}  &\multirow{4}{*}{NOTA \textcolor{darkgreen}{\ding{51}}}  &\multirow{4}{*}{NOTA \textcolor{darkgreen}{\ding{51}}}  &\multirow{4}{*}{P206 \textcolor{darkred}{\ding{55}}}  &\multirow{4}{*}{P206 \textcolor{darkred}{\ding{55}}}  &\multirow{4}{*}{P25 \textcolor{darkred}{\ding{55}}}\\
			\cline{2-8}
			&(7) He was a member of Adolph Rupp's `Fabulous Five'' University of Kentucky \textcolor{custompurple}{\textbf{basketball}} team, with Alex Groza, Wallace Jones, Cliff Barker, and \textcolor{customgreen}{\textbf{Kenny Rollins}}.
			&\multirow{4}{*}{\textcolor{red}{NOTA} (P641)}  &\multirow{4}{*}{NOTA \textcolor{darkgreen}{\ding{51}}}  &\multirow{4}{*}{NOTA \textcolor{darkgreen}{\ding{51}}}  &\multirow{4}{*}{P413 \textcolor{darkred}{\ding{55}}}  &\multirow{4}{*}{NOTA \textcolor{darkgreen}{\ding{51}}}  &\multirow{4}{*}{NOTA \textcolor{darkgreen}{\ding{51}}}\\
			\cline{2-8}
			&(8) \textcolor{customgreen}{\textbf{HD 37124}} c is an extrasolar planet approximately 108 light-years away in the constellation of \textcolor{custompurple}{\textbf{Taurus}}.  
			&\multirow{3}{*}{\textcolor{red}{NOTA} (P59)}  &\multirow{3}{*}{P206 \textcolor{darkred}{\ding{55}}}  &\multirow{3}{*}{NOTA \textcolor{darkgreen}{\ding{51}}}  &\multirow{3}{*}{P206
			\textcolor{darkred}{\ding{55}}}  &\multirow{3}{*}{P206 \textcolor{darkred}{\ding{55}}}  &\multirow{3}{*}{NOTA \textcolor{darkgreen}{\ding{51}}}\\
			\cline{2-8}
			&(9) \textcolor{customgreen}{\textbf{Warren Norris}} (born in St. John's, Newfoundland) is a Canadian professional \textcolor{custompurple}{\textbf{ice hockey}} centre who currently plays for EC KAC in the Austrian Hockey League.  
			&\multirow{4}{*}{\textcolor{red}{NOTA} (P641)}  &\multirow{4}{*}{NOTA \textcolor{darkgreen}{\ding{51}}}  &\multirow{4}{*}{P413 \textcolor{darkred}{\ding{55}}}  &\multirow{4}{*}{NOTA
			\textcolor{darkgreen}{\ding{51}}}  &\multirow{4}{*}{P413 \textcolor{darkred}{\ding{55}}}  &\multirow{4}{*}{NOTA \textcolor{darkgreen}{\ding{51}}}\\
			\cline{2-8}
			&(10) If mobilise, the NGVR would come under the operational command of the 8th Military District which was in the process of being raised under the command of \textcolor{custompurple}{Major General} \textcolor{customgreen}{Basil Morris}.  
			&\multirow{4}{*}{\textcolor{red}{NOTA} (P410)}  &\multirow{4}{*}{NOTA \textcolor{darkgreen}{\ding{51}}}  &\multirow{4}{*}{NOTA \textcolor{darkgreen}{\ding{51}}}  &\multirow{4}{*}{NOTA
			\textcolor{darkgreen}{\ding{51}}}  &\multirow{4}{*}{NOTA \textcolor{darkgreen}{\ding{51}}}  &\multirow{4}{*}{NOTA \textcolor{darkgreen}{\ding{51}}}\\
			\bottomrule
		\end{tabular}
	\end{adjustbox}
\end{table*}

\subsection{Case Study}

In order to intuitively demonstrate the effect of GPAM, we conduct a case study and artificially constructs a difficult and confusing 5-way-1-shot 0.5 meta tasks. For test requirements, we choose support and query set instances from five certain known classes, and select NOTA instances for query set from the remaining classes. And we choose BERT-PAIR \cite{DBLP:conf/emnlp/GaoHZLLSZ19} as a baseline model to compare it with the original GPAM and its three variant models: 1) $\text{GPAM}^{\#1}$ is the variant model that removes the semi-factual representation module; 2) $\text{GPAM}^{\#2}$ is the variant model that uses Euclidean distance instead of Mahalanobis distance as the metric; 3) $\text{GPAM}^{\#3}$ is the variant model that removes the margin strategy and only uses range for boundary division. We show some of the instances and the corresponding extraction results in a batch during the validation process in Tabel \ref{tab:my_label7}. Next, we analyze the results of the case study in detail.

Overall, GPAM performs best among the ten cases, correctly judging 90\% of the cases, followed by $\text{GPAM}^{\#3}$. For most known class query instances such as (1)-(3), all five models obtain correct results. For the query instance (4), sentence clearly implies that \textit{Chauncey Davis} is the \textit{defensive end}, which is a \textit{position played on team}. But BERT-PAIR and $\text{GPAM}^{\#2}$ fail to get the correct answer and misjudge it as the relation \textit{located in body of water}. For the query instance (5), which is considered one of the most challenging instances in this batch, \textit{Magas was the godfather of Serbian organized crime}, where \textit{Magas}, as the \textit{godfather}, was the mastermind of \textit{organized crime}, so the relation is \textit{main subject}. Only GPAM and $\text{GPAM}^{\#2}$ are correct, the other three models assume that the relation between entities is not any of the five and get the result NOTA. Instances (4) and (5) show that both the semi-factual representation and Mahalanobis distance metric strategies contribute to solving the few-shot problem of known classes.

For the unknown class query instances in Table \ref{tab:my_label7}, BERT-PAIR, GPAM and the variant $\text{GPAM}^{\#3}$ work relatively well in distinguishing NOTA classes, while $\text{GPAM}^{\#1}$ and $\text{GPAM}^{\#2}$ may misclassify a NOTA class as a known class multiple times. This phenomenon occurs because some NOTA class relations have similar semantics to known class relations, such as P641 (sport team) and P431 (position played on team), P59 (constellation position) and P206 (located in body of water). We select the instance (6) where both are wrong for detailed analysis, the sentence describes that the road stretches from the \textit{Pitt River Bridge} in the east to the \textit{Port Mann Bridge} in the west and runs right next to the \textit{Fraser River}. The correct relation \textit{NOTA} (P177), expresses the relation that the bridge \textit{Port Mann Bridge} spans the river Fraser River, instead of locating in body of water (P206) or other relationships. This shows that our complete GPAM is more accurate in identifying NOTA classes.

\section{Conclusion}
\label{Conclusion}

In this paper, we propose a model based on Gaussian prototype and adaptive margin named GPAM to solve few-shot relation extraction with NOTA task. The three core modules of GPAM are the semi-factual representation, the GMM-prototype learning and the decision boundary learning module. Besides, in decision boundary learning, we design a pseudo negative sampling strategy for NOTA scenarios to enhance the classification performance of the model. Experimental results on the FewRel dataset demonstrate that the performance of GPAM is better than that of the comparison methods, and ablation experiments convey the effectiveness of the designed modules and optimization strategies.
In the future, we hope to study the ability of LLMs on FsRE with NOTA task by combining semi-factual representation and adaptive margin.

\section*{Acknowledgment}
This work was supported by National Key Research and Development Program of China (2022YFC3303600), National Natural Science Foundation of China (62137002, 62293553, 62293554, 62176207, and 62192781), "LENOVO-XJTU" Intelligent Industry Joint Laboratory Project, Natural Science Basic Research Program of Shaanxi (2023-JC-YB-593), the Youth Innovation Team of Shaanxi Universities, XJTU Teaching Reform Research Project "Acquisition Learning Based on Knowledge Forest".

\bibliographystyle{elsarticle-num} 
\nocite{*} 
\bibliography{reference}

\newpage
\vspace{1em}

\begin{minipage}{0.2\textwidth}
	\includegraphics[width=1in,height=1.25in,clip,keepaspectratio]{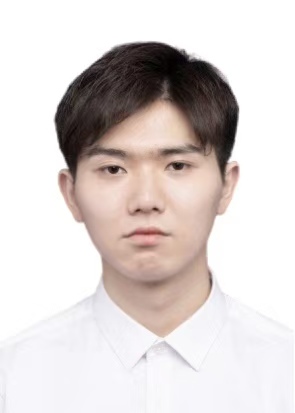}
\end{minipage}
\hspace{1em}
\begin{minipage}{0.72\textwidth}
	\textbf{Tianlin Guo} received the B.S. degree from Xi'an Jiaotong University, Xi'an, Shaanxi, China, in 2023, where he is currently pursuing the M.S. degree with the School of Computer Science and Technology. His research interests include information extraction and few-shot learning.
\end{minipage}

\vspace{1em}

\begin{minipage}{0.2\textwidth}
	\includegraphics[width=1in,height=1.25in,clip,keepaspectratio]{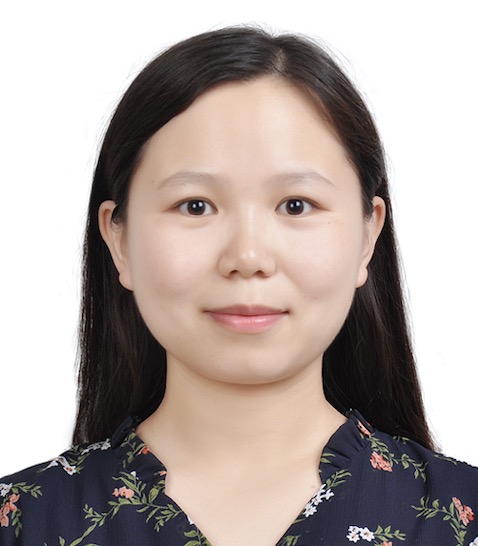}
\end{minipage}
\hspace{1em}
\begin{minipage}{0.72\textwidth}
	\textbf{Lingling Zhang} is currently an associate professor in computer science at Xi’an Jiaotong University. She received the PhD degree in Computing Science from Xi’an Jiaotong University in 2020. She was a visiting student with the School of Computer Science, Carnegie Mellon University, working with Prof. A. Hauptmann. Her research interests include cross-media information mining, computer vision, zero-shot learning, and few-shot learning.
\end{minipage}

\vspace{1em}

\begin{minipage}{0.2\textwidth}
	\includegraphics[width=1in,height=1.25in,clip,keepaspectratio]{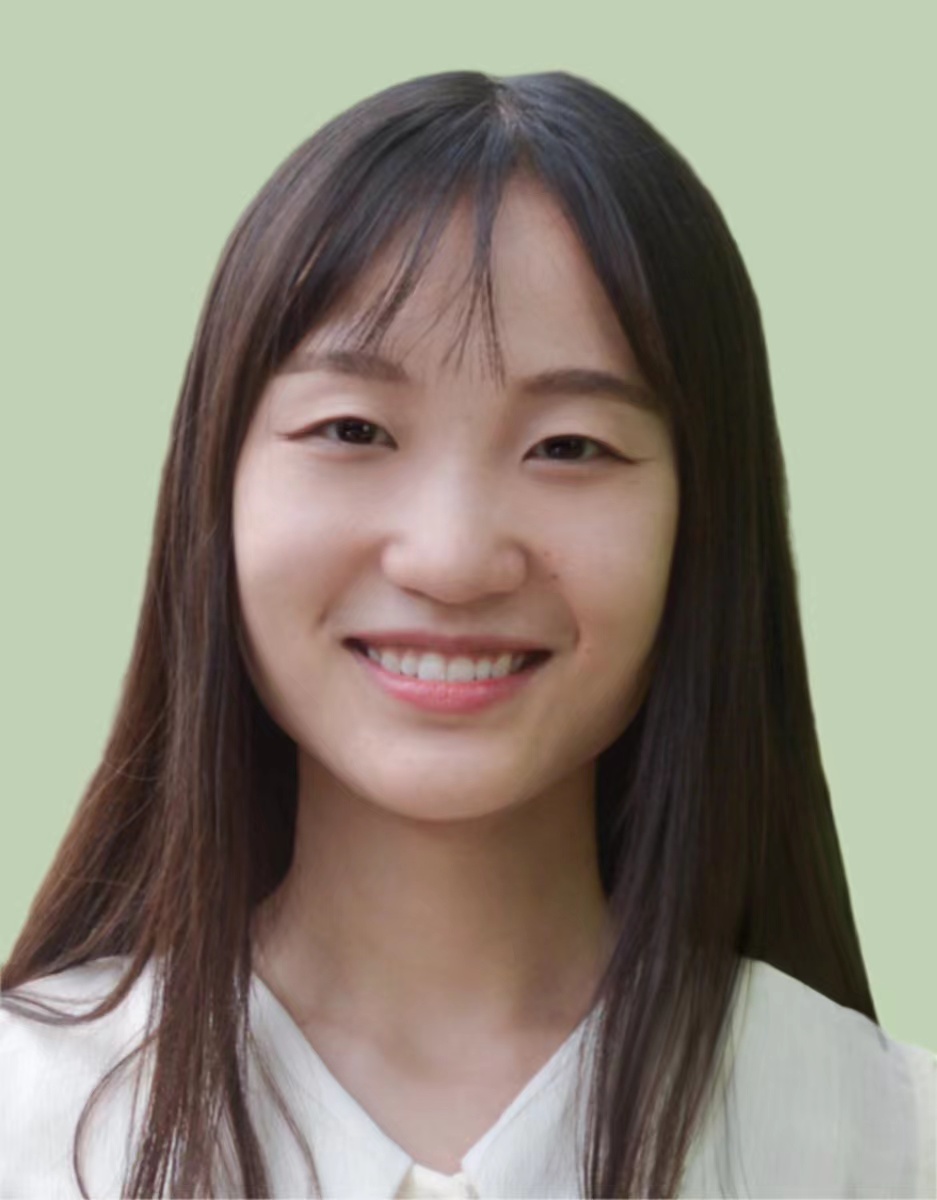}
\end{minipage}
\hspace{1em}
\begin{minipage}{0.72\textwidth}
	\textbf{Jiaxin Wang} received the B.S. and M.S. degrees in communication and information systems from Northwestern Polytechnical University, Shaanxi, China, in 2017 and 2020, respectively. She is currently pursuing the Ph.D. degree with the School of Computer Science and Technology, Xi’an Jiaotong University. Her research interests include natural language processing, large language models, and few-shot learning.
\end{minipage}

\vspace{1em}

\begin{minipage}{0.2\textwidth}
	\includegraphics[width=1in,height=1.25in,clip,keepaspectratio]{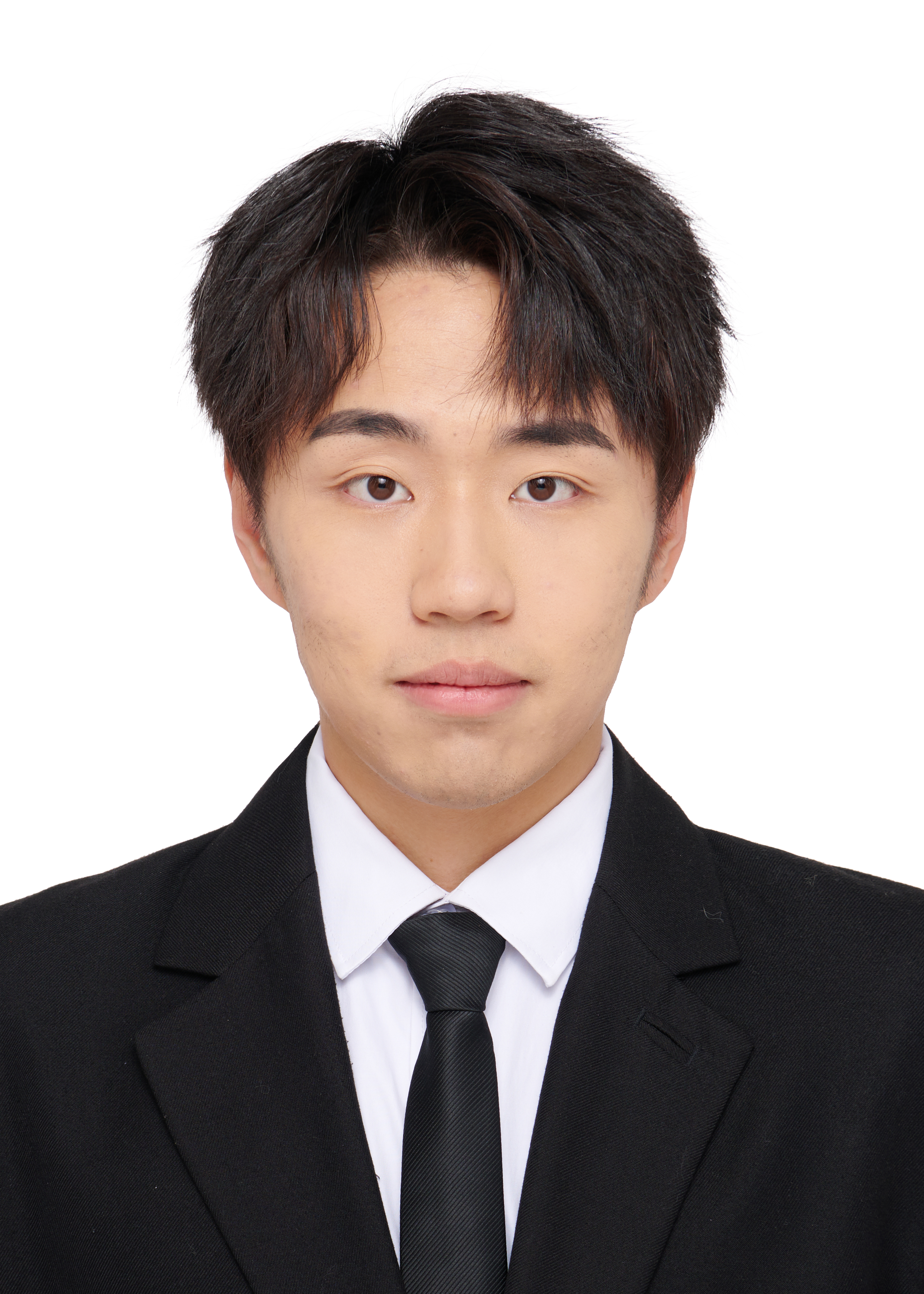}
\end{minipage}
\hspace{1em}
\begin{minipage}{0.72\textwidth}
	\textbf{Yunkuo Lei} received the B.S. degree from Xi'an Jiaotong University, Xi'an, Shaanxi, China, in 2024, where he is currently pursuing the M.S. degree with the School of Computer Science and Technology. His research interests include information extraction and spiking neural networks.
\end{minipage}

\vspace{1em}

\begin{minipage}{0.2\textwidth}
	\includegraphics[width=1in,height=1.25in,clip,keepaspectratio]{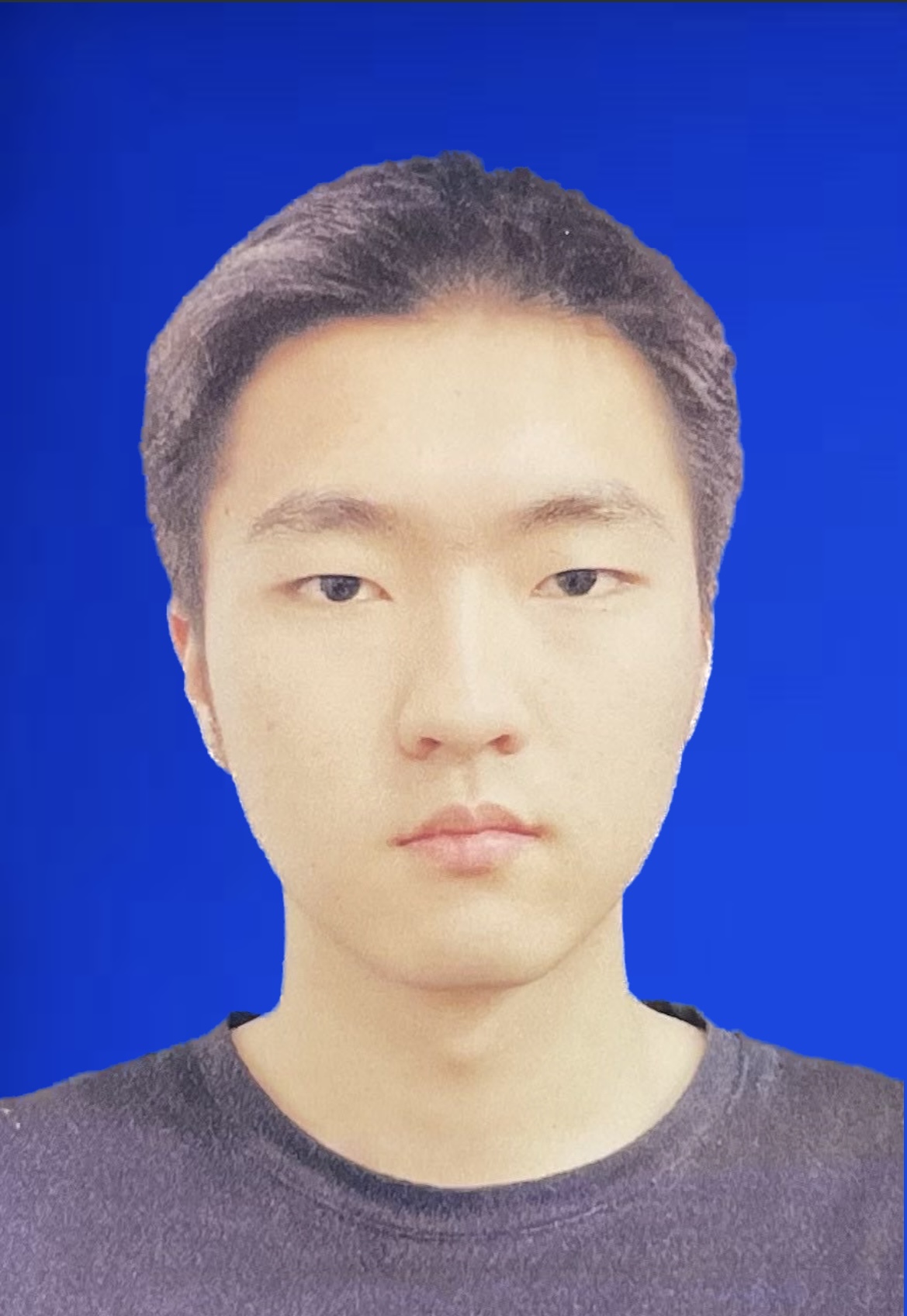}
\end{minipage}
\hspace{1em}
\begin{minipage}{0.72\textwidth}
	\textbf{Yifei Li} is currently working toward the Ph.D. degree in computer science at Xi'an Jiaotong University. He obtained his bachelor's degree from Xi'an Jiaotong University in 2022. His research interest is knowledge graphs and information extraction.
\end{minipage}

\vspace{1em}

\begin{minipage}{0.2\textwidth}
	\includegraphics[width=1in,height=1.25in,clip,keepaspectratio]{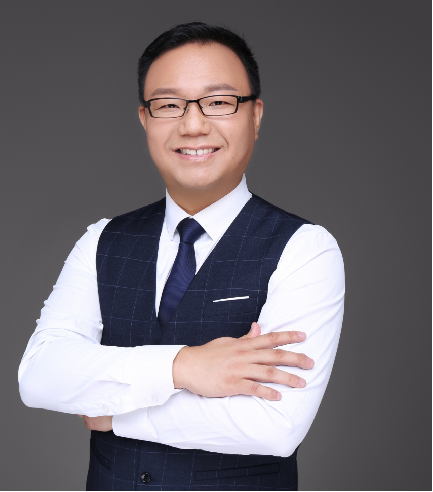}
\end{minipage}
\hspace{1em}
\begin{minipage}{0.72\textwidth}
	\textbf{Haofen Wang} is a professor with the College of Design and Innovation, Tongji University. He has taken charge of several national AI projects and published more than 100 related papers on top-tier conferences and journals. He has also served as deputy directors or chairs for several NGOs like CCF, CIPS and SCS.
\end{minipage}

\vspace{1em}

\begin{minipage}{0.2\textwidth}
	\includegraphics[width=1in,height=1.25in,clip,keepaspectratio]{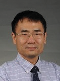}
\end{minipage}
\hspace{1em}
\begin{minipage}{0.72\textwidth}
	\textbf{Jun Liu} received the B.S., M.S., and Ph.D. degrees in computer science in 1995, 1998, and 2004 from Xi'an Jiaotong University, China. He is currently a professor in the School of Electronic and Information Engineering at Xi'an Jiaotong University. His research interests include text mining, data mining, intelligent network learning environments, and multimedia e-learning.
\end{minipage}

\end{document}